# Managing Uncertainty in LLM-Generated Procedural Knowledge for Virtual Laboratory Planning

Polychronis Karpodinis[1] and Dimitris Kalles[1][0000-0003-0364-5966]

[1] School of Science and Technology, Hellenic Open University, Patras 26335, Greece

**Abstract.** Educational virtual laboratories can make experimental training more scalable, adaptive, and accessible, especially when students have limited access to physical laboratory facilities. However, authoring new simulated laboratory procedures remains costly: educators must describe new equipment, define how instruments and materials interact, and specify valid procedural flows that can be executed or assessed inside the virtual environment. Large language models can assist in this authoring process by generating detailed experimental procedures, but their output should not be treated as directly executable plans. They may omit necessary actions, arrange steps in the wrong order, or produce instructions that are logically incorrect or incompatible with the laboratory equipment. This paper presents a prototype framework for managing uncertainty in LLM-generated procedural knowledge for virtual laboratory planning. The framework aims to reduce procedural uncertainty by using structured domain representations and uncertain LLM-generated state-transition samples to extract candidate procedural rules, transform them into explicit and inspectable constraints, and use them to repair uncertain procedural steps. Although the motivating domain refers to educational virtual laboratories, the underlying problem is more general: managing uncertain procedural knowledge for action planning in structured interactive environments. We illustrate the approach in a virtual laboratory domain involving laboratory instruments, containers, tools, and material-transfer actions.

**Keywords:** LLMs, uncertainty management, procedural knowledge, procedural planning, virtual laboratories, MDPs.

## 1 Introduction

Virtual laboratories offer students the opportunity to practice scientific procedures in simulated environments before or instead of performing them in a physical laboratory. This work is motivated by Onlabs [1], an interactive 3D virtual laboratory environment developed at the Hellenic Open University to support laboratory training in distance education, which offers predefined experimental scenarios, virtual instruments, instructional guidance, and assessment-oriented interaction tracking. Such environments are especially important in distance education, where students may have limited access to laboratory facilities, and can make experimental training more scalable and cost-effective.

However, virtual laboratories are often difficult to extend. Adding a new instrument, activity, or experimental procedure typically requires the collaboration of domain experts and software developers. In the Onlabs development process, a new activity begins with an analytical step-by-step description of the experimental scenario, written so that even users without prior biology experience can follow it. This is followed by video and photo capture, 3D modelling, virtual-lab implementation, and testing. Fig. 1 summarizes this activity-development pipeline.

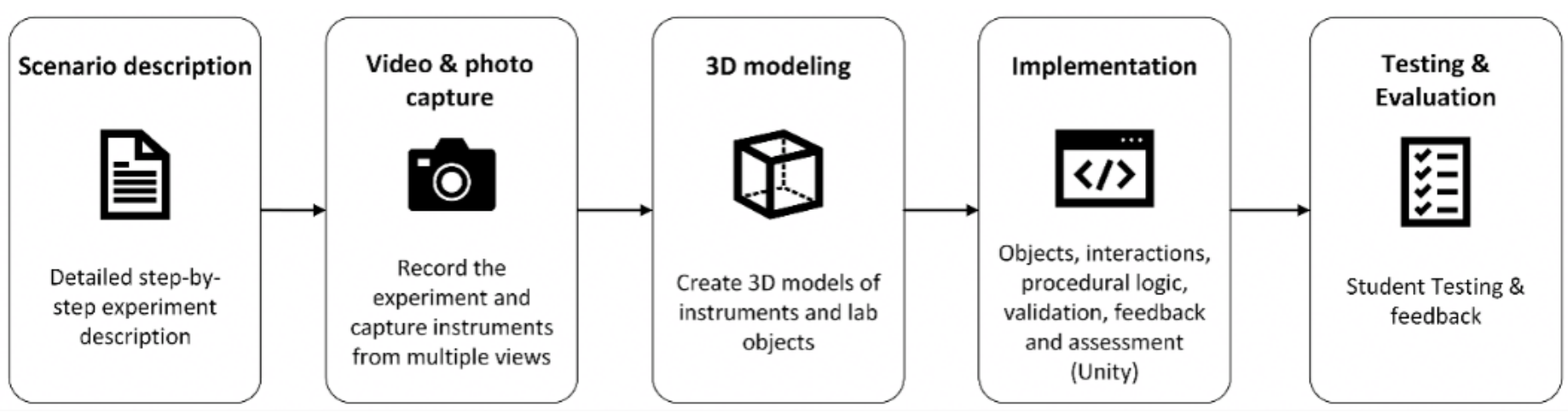

**Fig. 1.** Onlabs development pipeline.

The broader goal of this research is to reduce the effort required to develop and adapt virtual laboratory activities. Such environments can support laboratory education, especially when physical access is limited, but creating new simulations is expensive: each activity requires scenario design, object specification, interaction logic, implementation, and testing. The cost increases further when activities must scale across experiments or be adapted to different learning objectives, prior-knowledge levels, or personalization needs. Within this broader goal, we explore how LLMs can support authoring by generating draft experimental procedures and procedural knowledge about the laboratory domain. As shown in Fig. 1, this mainly supports the scenario-description stage, while also providing more explicit domain and constraint information for the implementation stage. However, even when prompts are grounded in available equipment, materials, and interactions, LLM-generated procedures remain uncertain: they may be incomplete, inconsistent, prompt-dependent, or plausible as text while still being operationally invalid, for example by using a scale before it is powered on, pouring from a closed bottle, or omitting a preparation step. In this paper, we therefore present a prototype pipeline for managing this uncertainty by structuring and evaluating generated procedural knowledge, extracting candidate rules, and using them to support procedure repair.

Earlier work in our project explored a hybrid approach in which an LLM generated a draft procedure, and a constraint-based layer refined it using manually specified domain rules. That approach demonstrated the promise of combining generative flexibility with structured validation, but it also exposed an important limitation: as the domain grows, manually encoding all procedural rules becomes difficult to scale. The present work therefore shifts emphasis from handcrafted validation rules toward extracting procedural knowledge from LLM-generated transition samples. The LLM is used not only to generate draft procedures, but also to support the discovery of candidate procedural constraints. Since the samples used to infer these constraints are themselves uncertain, the framework must manage uncertainty at the rule-extraction level to reduce uncertainty at the procedure level.

The pipeline begins with a structured model of the laboratory environment, including objects, components, states, actions, and interactions. From this representation, object-specific templates are generated and used to prompt the LLM to produce Markov Decision Process (MDP) transition samples, where each sample describes a state, an action, the resulting next state, and the LLM's plausibility judgment. These samples are treated as uncertain observations rather than ground-truth dynamics. They are aggregated into a tabular world model and analyzed to extract candidate preconditions, forbidden state values, and causal precedence rules. These rules can support missing-step insertion and provide ordering constraints for downstream repair of LLM-generated procedures. The repair stage assumes that an LLM-generated procedure may be broadly useful but locally flawed; therefore, the goal is to improve the draft rather than plan from scratch.

The contribution of the work presented in this paper is threefold:

a. It frames LLM-generated laboratory procedures and LLM-generated MDP transition samples as imperfect procedural knowledge requiring uncertainty management.
b. It proposes a pipeline that extracts candidate procedural rules from uncertain transition samples and turns them into explicit, inspectable constraints.
c. It formulates procedure improvement as constraint-guided repair of an approximately correct LLM-generated plan, rather than planning from scratch.

## 2 Related Work

This work is related to research on LLMs for planning, action reasoning, and uncertainty management in procedural domains. Recent work has shown that LLMs can generate plausible plans and action descriptions, but their outputs are often unreliable without explicit validation or grounding. In this context, Valmeekam et al. [2] developed the PlanBench benchmark to systematically evaluate LLMs on planning and reasoning-about-change tasks, showing that their performance remains limited when precise action ordering, preconditions, and state changes are required. Complementary approaches such as SayCan [3] and ReAct [4] explore how language models can be guided through external grounding, affordance estimates, environment interaction, or feedback. Together, these works suggest that LLM-generated action knowledge should be treated as provisional and subject to explicit validation rather than as directly reliable procedural knowledge.

These limitations connect LLM-based planning to the broader problem of uncertainty in structured knowledge representations. Alomair et al. [5] emphasize that uncertainty may arise from imperfect information, such as incomplete, imprecise, ambiguous, or inconsistent domain knowledge. This perspective is relevant when LLM-generated procedural content is not treated as ground truth, but as information that needs further structuring and evaluation. Neurosymbolic approaches address a related reliability problem from another direction. Bayless et al. [6] translate natural-language policies into formal logical artifacts and verify statements against them using symbolic reasoning. Such work reflects a broader trend toward combining LLMs with explicit, inspectable representations when reliability and auditability are important.

# 3 From Domain Model to Procedural Rules

Managing uncertainty begins by distinguishing what is explicitly specified from what is generated. The virtual laboratory is first described in a structured form, which provides a controlled vocabulary for the later LLM-based stages. Uncertainty enters when the LLM generates possible transitions and judges their plausibility; these uncertain samples are then aggregated and analyzed to extract candidate procedural rules.

## 3.1 Structured Representation of the Laboratory Domain

In the proposed uncertainty-management pipeline, the first stage is a structured representation of the virtual laboratory. The domain is represented as a JSON inventory of objects, components, object-level states, actions, and interactions. Objects may include instruments, containers, tools, and materials. Instruments are decomposed into components such as buttons, displays, platforms, receptors, or selectors. Each component may have its own actions and state space. The representation is encoded as structured JSON, but it can be constructed through a graphical interface in which a domain expert or educator defines laboratory objects, components, states, actions, and interactions. This supports the broader goal of making virtual-lab authoring accessible to non-programmers, while still producing a machine-readable representation that can be used for planning and rule extraction.

For example, an electronic scale can be represented through components such as a power button, display, weighing platform, and tare button, while interactions specify that aluminium foil can be placed on the weighing platform.

This structured representation provides semantic grounding for later stages. It constrains the vocabulary of states and actions, preventing later LLM outputs from being treated as unconstrained text. At the same time, the representation remains flexible enough to describe different instruments and interactions.

A key point is that not all state domains are fixed locally. Some state variables depend on interactions. For example, the possible values of a receptor component depend on which objects can be placed there. These dynamic state values are resolved when MDP templates are generated.

## 3.2 Object-Specific MDP Templates

An MDP, or Markov Decision Process, is a formal way to describe an environment in terms of states, actions, transitions, and rewards. Since MDPs are commonly used for decision-making in uncertain settings, they provide a natural formal basis for the uncertainty-management problem described above. In this work, the transition model is not assumed to be known in advance. Instead, object-specific templates, called MDP templates in this framework, act as structured dictionaries of the relevant state variables, values, and available actions for each laboratory object. These templates are later used to prompt the LLM to generate MDP transition samples, which are then aggregated into a tabular world model with empirical transition information and plausibility scores.

Each template is built around a particular instrument, tool, or container, but may include interaction-dependent context from other objects when needed. This avoids constructing one large state space for the entire laboratory, while still allowing the model to include external objects and state variables relevant to a given action. For example, the electronic-pipette template includes states of the pipette itself, such as whether it is powered on and which material it contains, but also relevant states of interacted objects, such as whether the double-distilled-water bottle is open or closed and which material is contained in the Erlenmeyer flask. These contextual states are included because the pipette can draw liquid from the bottle and pour it into the flask.

### 3.3 LLM-Generated MDP Transitions

After an MDP template has been created, it is used as structured input to an LLM. The model is asked to produce transitions of the form:

$$state, action, next_state, reward$$

The reward is binary in the current implementation: 1 means that the transition is judged plausible, while 0 means that it is judged implausible or invalid. Invalid or failing transitions are especially important. If the data contained only successful cases, it would be difficult to tell whether a condition is truly required for an action, or whether the model simply never explored cases where that condition was absent.

This generation step is also a source of uncertainty. The resulting data depends not only on the MDP template, but also on how the prompt presents the task, the available actions, and the expected form of judgment. Prompt design can therefore influence which transitions are generated and how their plausibility is assessed. For this reason, prompt dependence is treated as part of the uncertainty-management problem rather than as a minor implementation detail.

The generated transitions are therefore not ready-made rules; they are uncertain samples that must be aggregated and analyzed before they can support rule extraction. Their role is to provide material for the next stages of the pipeline, where repeated observations are aggregated and analyzed before candidate procedural rules are extracted.

## 4 World Model Construction

The generated transitions are aggregated into a tabular world model. The purpose of this step is to move from isolated LLM outputs to accumulated evidence about how actions behave in particular states. Transitions with the same state and action are grouped together, and the observed next states for each pair are recorded. For each state-action pair, the world model also computes a plausibility score from the generated rewards. This score is later used to separate likely valid cases from likely invalid or ambiguous ones.

This tabular form is deliberately simple and inspectable. It allows the system to reason about action validity using repeated observations rather than a single LLM generation. The world model does not yet decide what the rules are; instead, it organizes uncertain transition data into a form that can be analyzed to extract candidate preconditions, forbidden values, and causal dependencies.

In the current implementation, this preparation is mainly probability-oriented: repeated observations are summarized through counts, empirical next-state frequencies, and plausibility scores. This is not the only possible interpretation. Since the goal is to reason from incomplete and imperfect transition data, other uncertainty formalisms may also be relevant. For example, a possibilistic view could focus less on how frequently an outcome was observed and more on which transitions or preconditions remain plausible, which alternatives are ruled out, and how strongly a candidate rule is supported by the absence of counterexamples.

# 5 Preconditions and Causal Rules Extraction

## 5.1 Valid, Invalid and Ambiguous State-Action Entries

The first step in rule extraction is to classify each state-action entry according to its plausibility score. Entries above a validity threshold are classified as valid, entries below a lower threshold are classified as invalid, and entries in the middle are treated as ambiguous. Ambiguous entries are excluded from rule extraction because their evidence is not clear enough to support reliable preconditions. This reduces the risk of deriving misleading rules from mixed or insufficient evidence.

## 5.2 Required and Forbidden Conditions

After the valid, invalid, and ambiguous entries have been separated, the extractor analyzes each action separately. For every state variable, it checks which values appear in the successful cases and which values appear in the failing cases. The analysis is weighted by the amount of transition data behind each world-model entry, so a state-action pair observed many times has more influence than a state-action pair supported by only one generated transition.

We use simple notation to describe the extraction logic. Let $v$ denote a candidate value of a state variable for a given action, such as *power_button = on* or *cap = opened*. Let $v'$ denote an alternative value of the same state variable and let $\gamma$ be the confidence threshold used for rule extraction.

For required conditions, the intuition is simple: a value becomes a candidate precondition when it appears in most of the valid evidence for an action. We refer to this as valid support:

$$valid\ support(v) = \frac{valid\ evidence\ containing\ v}{total\ valid\ evidence\ for\ the\ action}$$

A value is treated as a candidate required condition when its valid support is high enough and it clearly dominates the alternative values of the same state variable:

$$\mathrm{required(v)} \Leftarrow \mathrm{valid\ support(v)} \geq \gamma \wedge \forall \mathrm{v}' \neq \mathrm{v}, \mathrm{valid\ support(v')} \leq 1 - \gamma$$

In words, this means that $v$ is not just frequent among successful cases; it is the main value associated with success for that state variable. For example, if almost all successful executions of a pipette draw action occur when the pipette is powered on and empty, these values become candidate requirements for that action.

Forbidden values are extracted more cautiously. A value is not treated as forbidden simply because it appears in failing cases, since the failure may be caused by another state value in the same entry. The extractor therefore first checks whether the value is absent, or almost absent, from valid evidence. It is then treated as forbidden only when there is additional support: either a one-value contrast, where changing only that value turns a valid case into an invalid one, or strong invalid-side support, used as a fallback when such contrastive evidence is not available:

$$invalid\ support(v) = \frac{invalid\ evidence\ containing\ v}{total\ invalid\ evidence\ for\ the\ action}$$

Thus, a value is treated as a candidate forbidden value when it is almost absent from valid evidence and either has one-value contrastive support for failure or has high invalid-side support:

$$forbidden(v) \Leftarrow valid\ support(v) \approx 0 \wedge (contrast(v) > 0 \vee invalid\ support(v) \geq \gamma)$$

Values that do not clearly satisfy the required or forbidden criteria remain neutral. This keeps the extracted rules closer to procedural requirements rather than simple correlations in the generated data.

### 5.3 Strong and Weak Preconditions

The extracted preconditions are not all equally certain. A precondition is strong when one value is required and the alternative values of the same variable are also identified as forbidden. In that case, the evidence supports both sides of the rule: the value appears necessary for success, and its alternatives appear incompatible with success. A precondition is weak when a value is strongly supported by valid evidence, but the available evidence is not sufficient to rule out all alternatives. This distinction makes remaining uncertainty explicit. In the current prototype, strong and weak preconditions are not yet treated differently during optimization; exploiting this distinction remains part of the work-in-progress direction.

### 5.4 From Preconditions to Causal Precedence Rules

The final rule-extraction step links necessary preconditions to producer actions. If an action $A$ requires state value $v$, the system searches the available MDP templates for actions that can produce $v$. This creates causal precedence rules of the form:

$$produces(B, v) \land requires(A, v) \Rightarrow B \prec A$$

In words, if action $B$ produces a state value required by action $A$, then $B$ should precede $A$.

For example, if pressing the tare button requires the scale power button to be on, and the action *power_button.set(value=on)* produces that state, then the system can infer a precedence relation requiring power-on action before the tare action.

## 6 Constraint-Guided Repair of Experimental Procedures

The extracted rules are used as candidate constraints for improving LLM-generated experimental procedures. Since the draft procedure may still contain useful high-level organization, the goal is not to plan from scratch, but to repair the existing sequence by reordering its steps while preserving as much of the original structure as possible. In the current prototype, this is formulated as a TSP-style permutation problem: each step is treated as a node, and the solution is a route through all steps. The draft is used both as a warm start and as a structural reference.

The objective combines four penalty terms:

$$\min \lambda_{\text{pos}} \sum_{i \in N} d_i \; + \; \lambda_{\text{edge}} \sum_{(i,j) \in E_0} (1 - x_{ij}) \; + \; \lambda_{\text{cluster}} \sum s_{k\ell}^{\text{cluster}} + \lambda_{\text{raw}} \sum s_{ij}^{\text{raw}}$$

where $d_i$ measures how far step $i$ moves from its original position, $x_{ij} = 1$ if step $j$ immediately follows step $i$in the repaired sequence and 0 otherwise, and $s_{ij}^{raw}$ and $s_{kl}^{cluster}$ measure violations of raw step-level precedence constraints and broader cluster-order constraints, respectively. Thus, $1 - x_{ij}$ penalizes draft adjacencies that are not preserved. The coefficients $\lambda_{pos}$, $\lambda_{edge}$, $\lambda_{cluster}$, and $\lambda_{raw}$ control the balance between preserving the draft and enforcing inferred constraints. The position and edge terms keep the repaired procedure close to the draft by limiting large step movements and preserving locally correct subsequences. The raw precedence term penalizes violations of extracted step-level rules, such as "open a bottle before drawing liquid," while the cluster term captures broader ordering preferences between groups of steps, such as placing liquid-transfer steps before solid-addition steps. In the current prototype, repair is included as a downstream use of the extracted rules, while the main focus of this paper remains the uncertainty-aware extraction of those rules.

# 7 Preliminary Evaluation

The preliminary evaluation has two parts. Section 7.1 examines the rule-extraction side of the pipeline, showing how a structured laboratory-domain description is transformed into MDP templates, generated MDP samples, world-model entries, candidate preconditions, and causal rules. Section 7.2 then evaluates how these rules are used in the constraint-guided repair stage to improve an LLM-generated procedure.

## 7.1 Rule Extraction Evaluation

We conducted a preliminary evaluation on a virtual-laboratory case study to examine whether the proposed pipeline can transform an abstract laboratory-domain description into explicit procedural constraints. The input was created through a GUI-based description tool and exported as JSON. The GUI allows the user to define the laboratory domain in terms of object categories, object instances, components, state variables, actions, and interactions between objects.

The case-study description includes three instruments — an electronic scale, a magnetic stirrer, and an electronic pipette — three material bottles for CuSO4, NaHCO3, and double-distilled water, and tools such as a spoon, aluminium foil, an Erlenmeyer flask, a magnetic stir bar, and a magnetic stick. These objects are described through components, states, and available actions. Interactions are also specified abstractly, as moving objects to receptor components or transferring material between objects. Thus, the GUI can describe relations such as placing aluminium foil on the scale, drawing liquid with the pipette, scooping material with a spoon, or pouring material into an Erlenmeyer flask, without manually encoding procedural rules such as “open the bottle before drawing liquid” or “turn on the scale before pressing tare”.

From this GUI/JSON description, the system generated object-specific MDP templates. For example, the electronic-pipette template includes states of the pipette itself, such as whether it is powered on and which material it contains. It also includes relevant states of interacted objects, such as the cap state of the double-distilled-water bottle and the material state of the Erlenmeyer flask. These contextual states are included because the pipette can draw liquid from the bottle and pour it into the flask. Accordingly, the template includes interaction actions such as drawing liquid from the bottle and pouring liquid into the flask, in addition to the pipette-specific control action for switching the pipette on or off.

In the next step, the MDP template is embedded in a prompt that asks the LLM to generate MDP samples, following the representation introduced in 3.1. The prompt constrains the generation to the state variables, action names, parameters, and value domains defined in the template, but still asks the LLM to use its general knowledge to produce both valid and invalid transitions and assign a reward. For each object-specific template, 250 samples were generated and then aggregated into the tabular world model.

Table 1 shows examples of generated MDP samples for the electronic pipette. Each example includes a concrete current state, action, reward, and next state, involving both pipette states and states of interacted objects. For readability, the tables use simplified labels, such as “bottle cap” or “flask material,” instead of the full internal JSON identifiers used by the implementation.

**Table 1.** Examples of generated MDP samples for electronic-pipette object.

| State | Action | Next State | Reward |
|---|---|---|---|
| power = on; pipette material = none; ddH2O bottle cap = opened; flask material = none | press draw button | pipette material = ddH2O; other unchanged | 1 |
| power = on; pipette material = none; ddH2O bottle cap = closed; flask material = none | press draw button | unchanged | 0 |
| power = on; pipette material = ddH2O; flask material = none | press pour button | pipette material = none; flask material = ddH2O | 1 |
| power = on; pipette material = none; flask material = none | press pour button | unchanged | 0 |
| power = on; pipette material = ddH2O; flask material = ddH2O | press pour button | pipette material = none; flask material = ddH2O | 0 |

As shown in Table 1, drawing from a closed bottle and pouring from an empty pipette are assigned reward 0, even though the state remains unchanged. In these cases, the LLM correctly judges the attempted action as invalid in that state. However, the last row shows an incorrect LLM judgment: pouring from a pipette that contains double-distilled water (ddH2O) is also assigned reward 0, although the action is valid and the resulting state is plausible. This clearly illustrates the uncertainty introduced by LLM’s judgments. The MDP samples generated are then grouped by identical state-action pairs into a tabular world model. For each state-action pair, the world model records the next states observed in the generated samples, how often each next state appeared, its probability, and the average reward assigned to that outcome. Table 2 shows examples of aggregated world-model entries for the electronic pipette.

**Table 2.** Examples of aggregated world-model entries for the electronic pipette.

| State-Action | Next State | Count | Prob. | Avg. Reward |
|---|---|---|---|---|
| power = off; pipette material = none; bottle cap = closed; flask material = none; action = power on | power = on; other unchanged | 23 | 1.00 | 1.00 |
| power = on; pipette material = none; bottle cap = opened; flask material = none; action = press draw button | pipette material = ddH2O; other unchanged | 27 | 1.00 | 1.00 |
| power = on; pipette material = none; bottle cap = closed; flask material = none; action = press draw button | unchanged | 9 | 1.00 | 0.00 |
| power = on; pipette material = ddH2O; bottle cap = opened; flask material = ddH2O; action = press pour button | pipette material = none; flask material unchanged | 7 | 0.35 | 0.86 |
| same state-action as previous row | unchanged | 13 | 0.65 | 0.00 |

The first three rows show cases where the generated evidence is consistent: the observed outcome has probability 1.00 and the average reward is either clearly valid or clearly invalid. The last two rows show a mixed case for the same state-action pair, where the LLM generated two different outcomes with different rewards. This indicates that uncertainty remains even after aggregation and the world model preserves the mixed evidence so that the later rule-extraction step can treat it more cautiously.

The world-model entries are then analyzed action by action to extract candidate preconditions. Values that are consistently associated with valid entries are marked as necessary conditions. Values that are absent, or almost absent, from valid entries and supported by invalid or contrastive evidence are marked as forbidden values. Table 3 shows the extracted candidate constraints for the electronic pipette.

**Table 3.** Candidate necessary and forbidden conditions extracted for the electronic pipette.

| Action | Candidate necessary condition | Candidate forbidden values |
|---|---|---|
| press draw button | bottle cap = opened;<br>pipette material = none;<br>power = on | bottle cap = closed;<br>pipette material = ddH2O;<br>power = off |
| press pour button | pipette material = ddH2O;<br>power = on | pipette material = none;<br>power = off |
| power on | power = off | — |
| power off | power = on | — |

Table 3 makes the extracted requirements explicit: drawing requires an open bottle, a powered-on pipette, and an empty pipette, while pouring requires the pipette to contain ddH2O and be powered on. Power-control actions are also captured: turning the pipette on requires it to be currently off, while turning it off requires it to be currently on. However, no candidate forbidden values are extracted for them, indicating remaining uncertainty that should be further managed, for example through additional generated evidence.

The final step links extracted preconditions to producer actions. If an action requires a state value and another action can produce that value, the system creates a causal precedence rule. Table 4 shows the resulting rules for the electronic pipette.

**Table 4.** Causal rules extracted for the electronic pipette.

| Action | Required Condition | Producer Action | Strength |
|---|---|---|---|
| press draw button | bottle cap = opened | open bottle cap | Strong |
| press draw button | pipette material = none | initial state / press pour button | Strong |
| press draw button | power = on | power on | Strong |
| press pour button | pipette material = ddH2O | press draw button | Strong |
| press pour button | power = on | power on | Strong |
| power off | power = on | power on | Weak |
| power on | power = off | initial state / power off | Weak |

Table 4 completes the pipette example by showing how candidate preconditions are converted into causal rules. In the producer column, “initial state” denotes a condition that may already hold before the procedure starts, while the listed action denotes a way to restore that condition later. The strong/weak distinction keeps uncertainty visible: in this case, the weak rules correspond to the control-action cases from Table 3, where a required condition was extracted but no candidate forbidden value was found.

Table 5 summarizes the causal-rule output for the six main objects in the virtual-laboratory case study: three material bottles and three instruments. The table does not list every atomic rule separately. Instead, it compresses producer-action links into readable sequence patterns. In the sequence column, A → B means that action A produces a condition required by action B. The + symbol means that several conditions or actions are required before the target action, without implying an order among them. The “required state / reset condition” column lists conditions that may hold initially or may need to be restored before repeating an action; for example, a spoon may be empty at the beginning, but before a second scoop it must be emptied by a previous pour action.

**Table 5.** Compact summary of causal-rule patterns for the bottle and instrument templates. S/W denotes strong/weak rules.

| Object | Causal sequence | Required state / Reset condition | Rules (S/W) |
|---|---|---|---|
| CuSO4 bottle | open cap → scoop CuSO4 → pour spoon to foil | spoon empty | 10 (7/3) |
| ddH2O bottle | {open cap + pipette on} → draw ddH2O → pour to flask; open cap → direct pour to flask | pipette empty | 11 (8/3) |
| NaHCO3 bottle | open cap → scoop NaHCO3 → pour spoon to foil | spoon empty | 7 (4/3) |
| Electronic pipette | {open bottle + pipette on} → draw ddH2O → pour to flask | pipette empty | 8 (6/2) |
| Electronic scale | power on → place foil on platform → tare | platform empty | 8 (4/4) |
| Magnetic stirrer | {place flask + insert stir bar + power on} → set nonzero rpm; nonzero rpm → zero rpm; insert stir bar → retrieve stir bar | platform empty | 31 (5/26) |

Table 5 shows that several extracted rules form meaningful procedural chains, such as opening a bottle before drawing or scooping, drawing liquid before pouring, and powering on an instrument before using it. At the same time, the strong/weak distinction keeps the remaining uncertainty visible. Weak rules usually indicate that a required condition was extracted, but the corresponding forbidden-value evidence was missing or incomplete. This is especially clear for the magnetic stirrer, where 26 of 31 rules are weak. The stirrer involves several interacting objects and coupled states, including the flask, stir bar, magnetic stick, platform, power state, and rpm state. This produces more residual uncertainty than simpler objects and suggests that this part of the domain needs further refinement, for example by generating additional targeted samples of contradictory cases.

The next subsection evaluates how these extracted causal rules can be used as repair constraints for an uncertain experimental procedure.

### 7.2 Constraint-Guided Repair Evaluation

The repair stage is still in progress and lies beyond the main focus of this paper. We include it here because it is the natural continuation of rule extraction: once candidate causal rules have been extracted, they can be used as ordering constraints to improve an LLM-generated procedure. The following results are therefore intended as an initial feasibility check of this downstream use of the extracted rules, rather than as a complete evaluation of the repair optimizer.

For this check, we started from a ground-truth experimental procedure and created a perturbed draft by introducing six local misorderings that resemble plausible LLM errors. Examples include transferring material before opening the corresponding container, closing a container too early, powering on the magnetic stirrer after using it, and powering off the stirrer before resetting the RPM value to zero. The repair model used causal constraints derived from the rule-extraction stage. In the current prototype, these rules were manually aligned with the concrete procedure steps before being passed to the optimizer; automating this mapping remains future work.

In this repair evaluation, only raw step-level precedence constraints were used; cluster-level constraints were not considered. We ran a small grid search over the repair weights controlling position preservation, edge preservation, and raw causal-precedence penalties. The best configuration was $\lambda_{pos} = 0.5$, $\lambda_{edge} = 1$, and $\lambda_{raw} = 2$. The reported metrics compare the perturbed draft and the repaired sequence against the ground truth from complementary perspectives: bigram and trigram overlap measure local ordering similarity, breakpoints count broken ground-truth adjacencies, LCS and Kendall's $\tau$ measure broader ordering agreement, displacement measures how far steps move from their ground-truth positions, and raw slack indicates whether the selected causal precedence constraints were satisfied.

The repaired sequence improves over the perturbed draft across all metrics: local overlap increases, breakpoints decrease, LCS improves from 24/30 to 28/30, and Kendall's $\tau$ rises from 0.853 to 0.954. The maximum displacement drops from 24 to 6, and zero raw slack indicates that the selected precedence constraints were satisfied. This suggests that the extracted rules can support targeted repair while preserving much of the draft structure.

**Table 6.** Draft and repaired sequence comparison against the ground truth.

| Sequence | Bigram | Trigram | Break-points | LCS | Kendall $\tau$ | Mean Δ | Max Δ | Raw slack |
|---|---|---|---|---|---|---|---|---|
| Draft | 0.483 | 0.321 | 15 | 24/30 | 0.853 | 1.733 | 24 | — |
| Repaired | 0.793 | 0.679 | 6 | 28/30 | 0.954 | 1.533 | 6 | 0.0 |

## 8 Conclusions and Future Work

The paper shows that a structured virtual-lab description can be transformed into inspectable procedural artifacts: generated MDP samples, world-model entries, candidate preconditions, and causal rules. The evaluation also illustrates why uncertainty management is needed: LLM judgments may be wrong, aggregated evidence may remain mixed, and some extracted rules remain weak. The repair results further show that these uncertain rules can still support targeted improvement of a draft procedure.

This work is still at an early stage. A first direction concerns uncertainty in rule extraction. Weak rules often arise when the generated evidence is incomplete, ambiguous, or insufficient to rule out alternative state values. Future work will examine whether additional targeted MDP samples can reduce this residual uncertainty, especially by testing contradictory or borderline cases that provide stronger evidence for required and forbidden conditions. Prompt design also remains important, since the generated samples depend on how the model is asked to produce valid and invalid cases, how states and actions are described, and how plausibility is judged. More broadly, we will investigate whether possibility theory can provide an alternative or a complement to the current count- and threshold-based treatment, since uncertainty in LLM-generated MDP samples is often epistemic rather than simply random.

Future work will also consider higher-level ordering constraints. Experimental procedures often have broader phase structure, such as preparation, weighing, mixing, stirring, and cleanup, which could be captured as cluster-level constraints and combined with raw precedence rules during repair.

**Acknowledgments.** ChatGPT was used to support language polishing and improve the clarity of the manuscript.

## References

1. Hellenic Open University, “Onlabs Virtual Laboratory,” [Online]. Available: http://onlabs.eap.gr/. [Accessed: May 22, 2026]
2. K. Valmeekam, M. Marquez, A. Olmo, S. Sreedharan, and S. Kambhampati, “PlanBench: An Extensible Benchmark for Evaluating Large Language Models on Planning and Reasoning about Change,” arXiv:2206.10498, 2023.
3. M. Ahn et al., “Do As I Can, Not As I Say: Grounding Language in Robotic Affordances,” arXiv:2204.01691, 2022.
4. S. Yao et al., “ReAct: Synergizing Reasoning and Acting in Language Models,” arXiv:2210.03629, 2022.
5. D. Alomair, R. Khedri, and W. MacCaull, “A Comprehensive Review of Information Uncertainty Modelling in Domain Ontologies,” ACM Computing Surveys, vol. 58, no. 10, Article 245, 2026.
6. S. Bayless et al., “A Neurosymbolic Approach to Natural Language Formalization and Verification,” arXiv:2511.09008, 2025.